\documentclass{article}

\usepackage[final]{nips_2017}

\usepackage[utf8]{inputenc} 
\usepackage[T1]{fontenc}    
\usepackage{hyperref}       
\usepackage{url}            
\usepackage{booktabs}       
\usepackage{amsfonts}       
\usepackage{nicefrac}       
\usepackage{microtype}      
\usepackage{amsmath}
\usepackage{csquotes}
\usepackage{graphicx}

\title{DANCin SEQ2SEQ: Fooling Text Classifiers 
with Adversarial Text Example Generation}

\author{
  Catherine Wong\\
  Department of Computer Science\\
  Stanford University\\
  \texttt{catwong@cs.stanford.edu} \\
}

\begin{document}

\maketitle

\begin{abstract}
Machine learning models are powerful but fallible. Generating adversarial examples -  inputs deliberately crafted to cause model misclassification or other errors - can yield important insight into model assumptions and vulnerabilities. Despite significant recent work on adversarial example generation targeting image classifiers, relatively little work exists exploring adversarial example generation for text classifiers; additionally, many existing adversarial example generation algorithms require full access to target model parameters, rendering them impractical for many real-world attacks. In this work, we introduce DANCin SEQ2SEQ, a GAN-inspired algorithm for adversarial text example generation targeting largely black-box text classifiers. We recast adversarial text example generation as a reinforcement learning problem, and demonstrate that our algorithm offers preliminary but promising steps towards generating semantically meaningful adversarial text examples in a real-world attack scenario.
\end{abstract}

\section{Introduction}
Machine learning models are powerful but fallible. Despite their surge in popularity for applications ranging from transportation to healthcare, a growing body of research demonstrates that many models are vulnerable to adversarial examples: inputs crafted to deliberately fool a targeted model into outputting an incorrect result or class \citep{szegedy2013intriguing}. These examples successfully fool even state-of-the-art deep networks - in fact, recent work suggests that the very expressiveness of neural networks renders them particularly vulnerable to certain adversarial attacks \citep{goodfellow2014explaining}. 

	Generating and defending against adversarial examples is essential for better model security: as machine learning models deployed towards increasingly complex and high-risk domains, holding machine learning models to the security standards of any other software is essential to ensuring user safety and trust. Adversarial examples can also yield broader insights into the targeted models themselves - much as optical illusions can guide research in human cognition, studying where and how models make mistakes can shed critical light on what a model does and does not actually know.
    
	In this paper, we propose a method to generate adversarial text examples that fool or increase the probability of misclassification in binary text classifiers. Text classifiers are now arguably some of the most commonly deployed machine learning models, used in high-impact domains ranging from spam classification to medical record analysis; however, many standard techniques used to generate adversarial examples have largely focused on image classifiers, and rely on gradient-based, visually imperceptible changes to existing images that do not easily apply to discrete and semantically meaningful text sequences. Here, we reframe adversarial text example generation as a reinforcement learning task and draw on GAN-inspired learning techniques to generate examples attacking a targeted classifier \footnote{A quick side note on terminology: confusingly, the word \textit{adversarial} has been used in literature to apply both to \textit{adversarial learning}, as in generative adversarial networks (GANs), and \textit{adversarial examples}, malicious examples crafted to cause deliberate model errors. This work applies a GAN-like framework to construct malicious text examples, so we will attempt to make the distinction clear whenever possible.}. This method allows for practical blackbox attacks - unlike many other approaches, we do not rely on access to the target model parameters - and early empirical results demonstrate that this method can generate modified text examples that increase the probability of misclassification, while preserving semantic similarity to the original text.

\section{Related Work}
Szegedy et. al first described the vulnerability of deep neural networks to adversarial image examples, and proposed a method to find these examples by searching over visually imperceptible perturbations to existing images within the transformed image space \citep{szegedy2013intriguing}. Goodfellow et. al later demonstrated the fast gradient sign method, a much cheaper method to generate adversarial image examples by exploiting the linear behavior of neural networks \citep{goodfellow2014explaining}. Recent work has demonstrated that these adversarial image examples are alarmingly robust - many continue to cause model errors even when downsampled or printed - and therefore practical for real world attacks \citep{athalye2017synthesizing}.

Notably, these approaches calculate malicious perturbations based on model gradients, and require full access to the target model parameters to construct adversarial image examples. The high-dimensionality of the image input space poses a challenge for blackbox attacks. However, Papernot et. al. have demonstrated that many adversarial images are also robustly transferable across many network architectures, allowing gradient-based approaches used to target known networks to be used against other similar but unknown models \citep{papernot2017practical}.

Much less work exists, however, addressing adversarial example generation for text-based classifiers. Jia and Liang demonstrated that networks trained for more difficult tasks, such as question answering, can be easily fooled by introducing spurious, distracting sentences into text, but these results do not transfer obviously to simpler text classification tasks \citep{jia2017adversarial}. Techniques used for image example generation, however, do not directly apply in the text domain. An analogous approach would require modifying an input text example to cause misclassification while preserving semantics, but the discrete nature of text makes finding small but “similar”, let alone “imperceptible”, perturbations challenging. Caswell et. al. explore methods to replace words in an input example with their nearest neighbors in the transformed input space, but with mixed to unsuccessful results \citep{caswellexploring}.

\section{DANCin SEQ2SEQ: Adversarial Text Example Generation}

This section describes \textit{Dueling Adversarial Neural Classification in SEQ2SEQ models}, or \textit{DANCin SEQ2SEQ}, a GAN-like training algorithm to generate adversarial text examples. First, we introduce \textit{adversarial REINFORCE}, a policy gradient method first described by Li et. al that adapts the GAN framework for discrete text sequence generation \citep{li2017adversarial}. We then describe how this same idea can be modified for adversarial \textit{example} generation, targeting a black-box binary text classifier.

\subsection{Adversarial REINFORCE for Text Generation}
The idea of generative adversarial networks cannot be directly applied for text generation; unlike in images, text sequences are discrete, which makes the discriminator error hard to backpropagate to the generator.

The \textit{adversarial REINFORCE} algorithm uses policy gradient methods to adapt the GAN objective function for text generation \citep{li2017adversarial}. Li et. al. describe a formulation specifically intended for dialogue response generation, where a model is given a dialogue history $x$ consisting of a series of previous dialogue utterances, and must generate a text sequence response $y = {y_1, y_2, ...y_T}$. The algorithm reframes text sequence generation as a sequence of actions taken according to a policy defined by a recurrent neural network.

\begin{figure}[h]
\label{adversarial-reinforce}
  \centering
  \includegraphics[width=300pt]{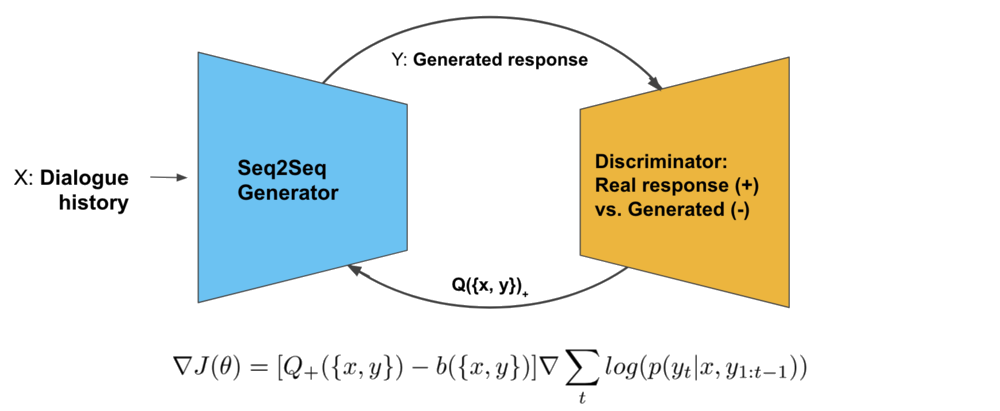}
  \caption{The Adversarial REINFORCE framework and policy gradient formulation.
}
\end{figure}

As with the standard GAN formulation, the algorithm consists of a generative model $G$ and a discriminative model $D$ [Figure 1]. The generative model $G$ takes a form similar to SEQ2SEQ models, and defines the policy to generate the response $y$ given the input dialogue history $x$. The discriminative model $D$ is a binary classifier that takes as input a sequence of dialogue utterances $\{x, y\}$, and outputs the probability that the input dialogue episode is machine-generated ($Q_-\{x,y\}$) or a real, human-generated dialogue episode ($Q_+\{x,y\}$).

The algorithm uses policy gradient training to encourage the generator to produce outputs that cannot be distinguished from human responses. The generator objective function attempts to maximize the expected reward of generated dialogue sequences, where the discriminator score is used as a reward in the REINFORCE formulation:

\begin{equation}
J(\theta)=\mathbb{E}_{y\sim p(y|x)}[Q_+ (\{x,y\}) | \theta]
\end{equation}

Given each input dialogue history $x$, the generator produces a generated response $y$ by sampling from the policy, and the pair $\{x,y\}$ is fed to the discriminator. The generator is then updated using a gradient approximated using the likelihood ratio trick, where $\pi$ is the probability of the generated responses, and $b(\{x,y\})$ is an unbiased baseline value to reduce the estimate variance:

\begin{equation}
\begin{split}
\nabla J(\theta) \approx \\
[Q_+(\{x,y\}) - b(\{x,y\})]\nabla log \ \pi (y | x) = \\
[Q_+(\{x,y\}) - b(\{x,y\})] \nabla \sum _t log \ p(y_t | x; y_{1:t-1})
\end{split}
\end{equation}

Again, as with standard GAN training, the discriminator is simultaneously updated with the human generated dialogue $\{x, y_{human}\}$ containing the dialogue history as a positive example, and the machine-generated dialogue $\{x, y_{generated}\}$ as a negative example.

\subsection{DANCin SEQ2SEQ}
We now describe a proposed algorithm, DANCin SEQ2SEQ, that draws on the REINFORCE formulation described above to instead generate adversarial \textit{examples} intended to fool a target binary text classifier model.

\subsubsection{REINFORCE for Adversarial Example Generation}
Specifically, we consider the problem formulated as follows: the target is a binary text classifier trained to discriminate between positively and negatively labeled text examples from a given dataset. Now, given a text example $x$ whose true label is positive, the adversarial model needs to generate a rewritten example $x'={x_1, x_2, ...x_T}$, where $x'$ should preserve semantic similarity to $x$, but fool the target model into outputting the incorrect negative label. 

We cast this problem as a reinforcement learning task, and propose a simple modification to the GAN-like REINFORCE formulation described in 3.1 to allow for adversarial example generation. In our formulation, the generative model $G$ defines the policy that takes the original positively-labeled input example $x$, and attempts to generate a rewritten adversarial example $x'$. 

We then use the \textit{target binary text classifier model} as the discriminator $D$, which receives the adversarial example $x'$ as input and outputs the probability that $x'$ is a positively-labeled example ($Q_+(x')$) or a negatively-labeled example ($Q_-(x')$). Notably, unlike other adversarial example generation algorithms, this formulation does not require any knowledge of the target model parameters, allowing the algorithm to attack largely black-box models with access only to the target model \textit{confidences} of its classifications.

As our goal is to encourage the generative model to perturb the positively-labeled input example so that the discriminator assigns an incorrect, negative label to the rewritten example x’, we can now simply modify the REINFORCE algorithm described in 3.1 so that the generator objective function rewards sequences that fool the discriminator:

\begin{equation}
J(\theta) = \mathbb{E}_{y \sim p(y|x)} [Q_- (\{x'\}) | \theta]
\end{equation}

Note that unlike the algorithm in 3.1, the target model D is not itself updated after each episode. Instead, this represents an attack where we learn to fool a targeted discriminator by interacting with it repeatedly, but where that target model itself remains fixed.

\subsubsection{Preserving Semantic Similarity with an Impartial Judge}

Importantly, the objective function above is not sufficient to capture the full adversarial text example generation problem: in addition to rewarding the generator for rewriting a positively-labeled input example $x$ so that the resulting text $x'$ is classified oppositely by the discriminator, we must also ensure that the rewritten example $x'$ remains semantically similar to the original, so that a human would still assign the original, positive label to $x'$. Intuitively, this is similar in spirit to the adversarial image generation problem, where algorithms attempt to perturb an input $x$ so that it remains visually close to the original, while still receiving a different classification from the targeted image classifier.

As discussed above, the discrete semantics of text pose a challenge to applying traditional adversarial example techniques, which generate “similar” or visually “close” perturbations using gradient-based methods. However, casting adversarial text example generation as a reinforcement learning task suggests a solution: we can update the generator objective function to reward sampled adversarial text examples that not only fool the target classifier, but also preserve semantic similarity to the original example. Intuitively, this approach also allows us to control the tradeoff between the adversarial nature and semantic similarity of the generated examples - that is, how much we care that our rewritten messages remain similar to the originals, versus simply leading the discriminator to assign an opposite label.

To reward semantic similarity between the original and rewritten examples, we introduce a third component into training, the \textit{impartial judge} J. Specifically, we use a judge J represented by a SEQ2SEQ autoencoder trained on a broad dataset of text examples from the same domain as the target classifier. We can then measure the similarity between two text examples based on the similarity of their vector representations when encoded by the impartial judge. Intuitively, adversarial examples attack target model overfitting to its own objective function - the trained model has learned an overly “narrow” view of the world by attempting to learn a binary classification function to separate its labeled training inputs. Therefore, the autoencoder judge trained without the goal of binary classification should be able to distinguish more impartially between true semantic similarity within the same text domain.

\begin{figure}[h]
\label{adversarial-reinforce}
  \centering
  \includegraphics[width=300pt]{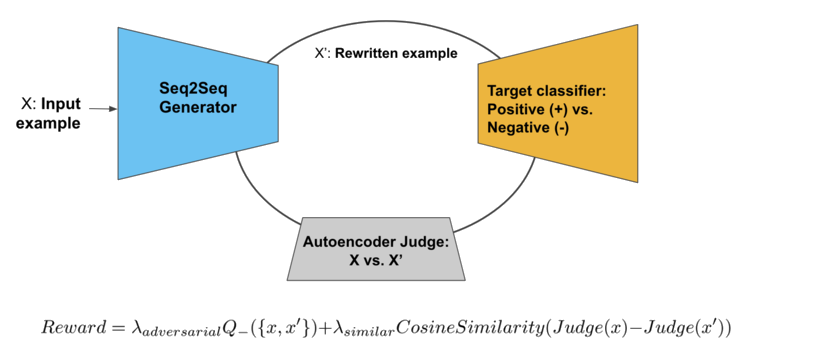}
  \caption{The DANCin SEQ2SEQ framework and updated reward function.
}
\end{figure}

We can now present the full DANCin SEQ2SEQ objective formulation [Figure 2], which rewards both misclassification by the target classifier and similarity between the original example $x$ and the rewritten example $x'$:

\begin{equation}
\begin{split}
J(\theta) = \\
\mathbb{E}_{y \sim p(y|x)} [\lambda _ {adversarial}Q_-(\{x'\}) + \lambda _ {similar} CosineSimilarity(Judge(x), Judge(x') | \theta]
\end{split}
\end{equation}

where $Judge(x)$ is the encoded vector representation of $x$ by the impartial judge, and $\lambda _ {adversarial}$ and $\lambda _ {similar}$ are hyperparameters controlling the weight placed on fooling the target classifier and preserving semantic similarity, respectively.

\section{Evaluation}
\subsection{Task}
We evaluate the DANCin SEQ2SEQ framework on the Enron Spam dataset, which consists of 16,545 labeled ham (non-spam) messages, and 17,171 labeled spam messages drawn from the released Enron corporation emails \citep{Enron}.

Specifically, we consider the task of attacking a fixed target spam-ham binary text classifier D, by generating adversarial examples x’ based on unseen input spam message examples x. Our generator model should attempt to rewrite the input spam examples to increase the probability that they are classified as \textit{not spam} by the target classifier, while still preserving human semantic similarity to the original spam messages. We choose this task to represent a realistic domain for adversarial example generation attacks, where misclassification holds real-world negative consequences. While the algorithm requires access to target classifier confidences in its classifications, the ease of gathering realistic training data also means that an attacker could reasonably employ a strategy similar to Papernot et. al’s practical black-box attack, in which adversarial training examples generated against a known model are then transferred to attack an actual, unknown target \citep{papernot2017practical}.

\subsection{Experimental Details}
\textit{\textbf{Dataset preprocessing.}} As token-level SEQ2SEQ models incur significant computational training costs in proportion to input vocabulary size and sequence length, for these demonstration experiments, we preprocess the dataset to restrict both. 

In particular, we first divide the full, randomly shuffled Enron Spam dataset into training, validation, and test dataset splits following a roughly 80-10-10 ratio, resulting in a training dataset consisting of {13,236 ham, 13,736 spam}, a validation dataset consisting of {1,654 ham, 1,717 spam}, and a test dataset consisting of {1,655 ham, 1,718 spam} messages. These dataset splits, along with all other dataset preprocessing code, are publicly available on Github \citep{github}.

To reduce sequence length, all text examples were lowercased, tokenized on white space, and then normalized to a fixed length of 30 tokens, by truncating longer examples and padding shorter examples with an introduced <PAD> token.

To reduce vocabulary size, we restricted the text example vocabulary to the set consisting of the 3,000 most frequent tokens in each of the spam and ham classes within the training dataset, resulting in a total vocabulary of 4,628 distinct tokens. All examples were then preprocessed according to this fixed vocabulary, and out of vocabulary tokens were simply replaced with an introduced <UNK> token.

Additionally, start and end of sequence tokens were pre and postpended to all examples.

\begin{figure}[h]
\label{adversarial-reinforce}
  \centering
  \includegraphics[width=\textwidth]{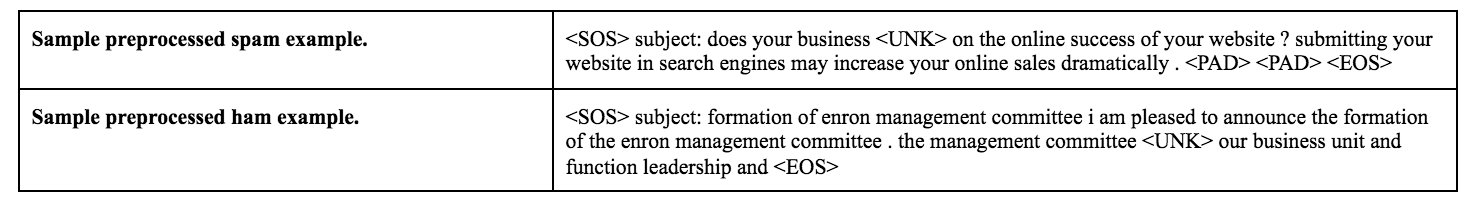}
  \caption{Sample spam and ham examples, after sentence length normalization and vocabulary size restriction.}
\end{figure}

\textit{\textbf{Model details.}}
As described, the DANCin SEQ2SEQ framework consists of three components: the adversarial example generator G, the target binary classifier D, and the impartial judge J. 

As a target binary classifier D, we use a Multinomial Naive Bayes model trained on the preprocessed training spam-ham dataset with Laplace smoothing parameter 1.0. Because we focus on designing practical, real-world attacks, we chose this model architecture after consulting NLP domain experts to best represent the actual model architectures most frequently used for real-world spam/ham classification. However, while the limited expressivity of a Naive Bayes model may in fact offer some robustness advantages to adversarial examples, as discussed earlier, we also recognize that the model simplicity may handicap its ability to recognize semantically similar adversarial examples; we therefore invite further experimentation on other target model architectures, such as RNN-based models, for future exploration. The resulting trained model D achieves 96.0\% validation accuracy with a 99.4\% AUC-ROC score, indicating that the model can in fact accurately distinguish spam and ham messages even after dataset preprocessing.

As an impartial judge J, we use an attentional SEQ2SEQ autoencoder, consisting of a 2-layer bidirectional LSTM encoder and a single layer LSTM decoder with source and target embedding dimensions 256, and source and target LSTM hidden dimensions 1024. The autoencoder was pretrained on the Enron spam-ham training dataset for 1000 epochs using a learning rate of 0.0002, Adam optimization over the cross entropy loss between the encoded and decoded vectors, and batch-size 50. Then, during DANCin SEQ2SEQ training, original and rewritten input examples were then encoded using the pretrained encoder.

As an adversarial example generator G, we use the same SEQ2SEQ architecture as the impartial judge - in fact, we initialize the model weights using the pretrained autoencoder weights on the training dataset. During DANCin SEQ2SEQ training, we use the modified objective function described in 3.2.2, with an unbiased baseline calculated as an exponentially decaying running average over the rewards with baseline weight 0.99 for the previous average at each timestep. We train using Adam optimization with a learning rate of 0.0002 and batch size of 10.

All fully-runnable model implementations, including trained model checkpoints, are publicly available on GitHub \citep{github}.

\textit{\textbf{Training details.}} All models were implemented in Python: we draw on the default Scikit-Learn Multinomial Naive Bayes implementation for the target model \citep{scikit}, and modify an existing PyTorch SEQ2SEQ implementation for the impartial judge and generator \citep{seq}. All models were trained on a Titan X GPU.

\subsection{Results}
\subsubsection{Full Unseen Spam Dataset Training}
\begin{figure}[h]
\label{adversarial-reinforce}
  \centering
  \includegraphics[width=\textwidth]{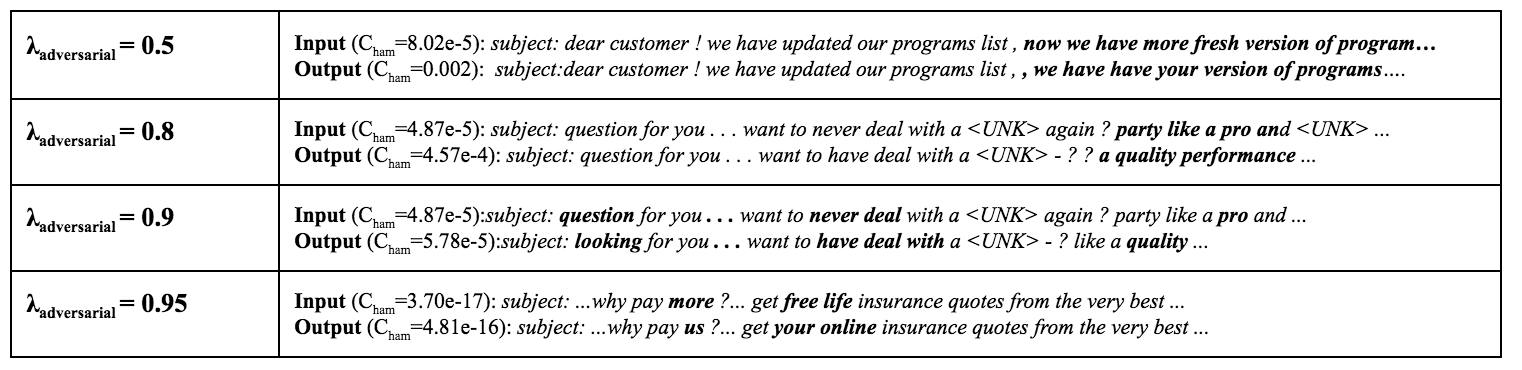}
  \caption{Excerpts from sampled original spam inputs and rewritten adversarial outputs after training on (n=1717) unseen spam examples.}
\end{figure}

We first train on n=1717 randomly sampled spam examples, previously unseen by the objective judge and the targeted classifier, using $\lambda _{adversarial}$ = \{0.5, 0.8, 0.9, 0.95\}, where $\lambda _ {aimilar}$ is set as $1 - \lambda _ {adversarial}$ . 

Results showing sample original spam examples and the generated adversarial rewritten examples are shown in Figure 4. From a subjective evaluation of sampled inputs and generated outputs during training, we observe that the generator does in fact appear to learn a number of intriguing, interpretable strategies for generating adversarial text examples:

\begin{enumerate}
\item \textit{\textbf{Semantically similar token replacement:}} perhaps most promisingly, we observe that the generator does in fact learn to replace tokens highly correlated with the spam label, such as mail, medication, software, and (somewhat unfortunately for humans deluged by spam) unsubscribe with semantically similar, but less spam-like, tokens. Two examples from training using $\lambda _{adversarial}$ =0.5 are given below, along with the discriminator log-probability confidence in the targeted non-spam label:
\begin{displayquote}
\textbf{Input} ($C_{ham}=3.78e{\text -}14$): \textit{subject: fwd...\textbf{look younger + more energy + lose weight} in three...}

\textbf{Generated adversarial output} ($C_{ham}=3.80e{\text -}10$): \textit{subject: fwd...my \textbf{age l i ve a long times weight weeks} in in best...}\\

\textbf{Input} ($C_{ham}=0.99$): \textit{subject: \textbf{smart spam} control...}

\textbf{Generated adversarial output}($C_{ham}=0.99$): \textit{subject: \textbf{enjoy internet} control…}
\end{displayquote}

\item \textit{\textbf{Deletion of spam-associated tokens:}} we also observe that the generator frequently learns to simply remove highly spam-correlated tokens, replacing them instead with more neutral tokens such as the <UNK> token or punctuation. Two examples of this are given below, again from training using  $\lambda _{adversarial}$=0.5. Notably, in the second example, the generator also adds the token \textit{international}, more closely associated with ham data from the Enron dataset.

\begin{displayquote}
\textbf{Input} ($C_{ham}=5.0e{\text -}7$): \textit{subject: professional advertising dear projecthoneypot @ projecthoneypot . org : we offer e - \textbf{mail} marketing with \textbf{best} services...}

\textbf{Generated adversarial output} ($C_{ham}=1.1e{\text -}6$): \textit{subject: professional advertising dear projecthoneypot @ 
projecthoneypot . . : we offer e - \textbf{<UNK>} marketing with \textbf{<UNK>} services
}\\

\textbf{Input} ($C_{ham}=3.3e{\text -}9$): \textit{subject: don , t <UNK> this , you have won a \textbf{prize} ! ! ! from : the desk of the managing 
director...}

\textbf{Generated adversarial output} ($C_{ham}=1.9e{\text -}7$): \textit{subject: don , t <UNK> <UNK> , you have won a \textbf{! ! ! !} from : <UNK> 
\textbf{international} of the managing director...
}
\end{displayquote}

\end{enumerate}
Figure 4 showcases additional examples demonstrating semantically-related token substitutions that increase the discriminator probability of misclassification. Notably, while these examples do increase the discriminator confidence in the incorrect non-spam class, they do not surpass the standard $P_{ham}=0.5$ threshold necessary for actual misclassification. However, the results are preliminary but promising, and suggest that the DANCin SEQ2SEQ formulation does in fact offer an approach to learn semantically meaningful adversarial perturbations against a largely black-box target model.

However, we also find that, as with the original adversarial REINFORCE algorithm for dialogue generation, GAN-style training is highly unstable; in all experiments, we observed a significant deterioration in output quality and ultimately complete generator mode collapse within 1-4 epochs of training, making more robust empirical evaluation on a dataset infeasible given time and computational resource restraints. In particular, we identify three primary failure cases in adversarial example generation: 

\begin{enumerate}
\item \textbf{Semantic divergence:} especially at higher values of $\lambda _ {adversarial}$, we find that the generator begins to produce rewritten examples that are no longer semantically similar to the original message, such as:
\begin{displayquote}
\textbf{Input}: \textit{subject: winning notifications ! ! ! easy way lottery agency number…}\\
\textbf{Output}: \textit{subject: winning winning ! ! ! our teens health health promotions...}
\end{displayquote}

\item \textbf{Reward-hacking:} especially as training progresses, the generator also tends to discover that it can also more easily gain a large reward from examples that receive high not-spam classification scores by exploiting the bag-of-words nature of the Naive Bayes target classifier. These examples are composed entirely of discovered tokens associated strongly with ham messages and read like a comical mishmash of “Enronese”; while they clearly do not preserve semantic similarity, the high discriminator confidence in the target label overrides the penalty for semantic dissimilarity. 
\begin{displayquote}
\textbf{Output}: \textit{... consulting hourahead calculation terminated agenda ebiz bro anderson anderson tiger fda fda bids...}
\end{displayquote}

\item \textbf{Mode collapse:} ultimately, as discussed above, the instability of GAN-style training resulted in full generator mode collapse for all training regimens, even after a preliminary search over learning rate and other hyperparameter adjustments. This same instability was also described in the related adversarial REINFORCE work, which suggests that mode collapse may appear due to reward sparsity in training - that is, the generator learns that it has stumbled into a low-reward domain and receives repeated negative feedback, but is unclear on how to correct this, ultimately leading to full divergence in most basic adversarial training regimens.
This mode collapse manifests in outputs composed of repetitive, low-meaning tokens.
\begin{displayquote}
\textbf{Output}: \textit{subject: ! ! ! ! ) ) ) ) ) ) ) ) ) ) ) ) ) ) ) ) ) ) ) ) ) ) ) ) ) )...
}
\end{displayquote}
 
\end{enumerate}

\subsubsection{Low-Confidence Spam Dataset Training}
\begin{figure}[h]
\label{adversarial-reinforce}
  \centering
  \includegraphics[width=\textwidth]{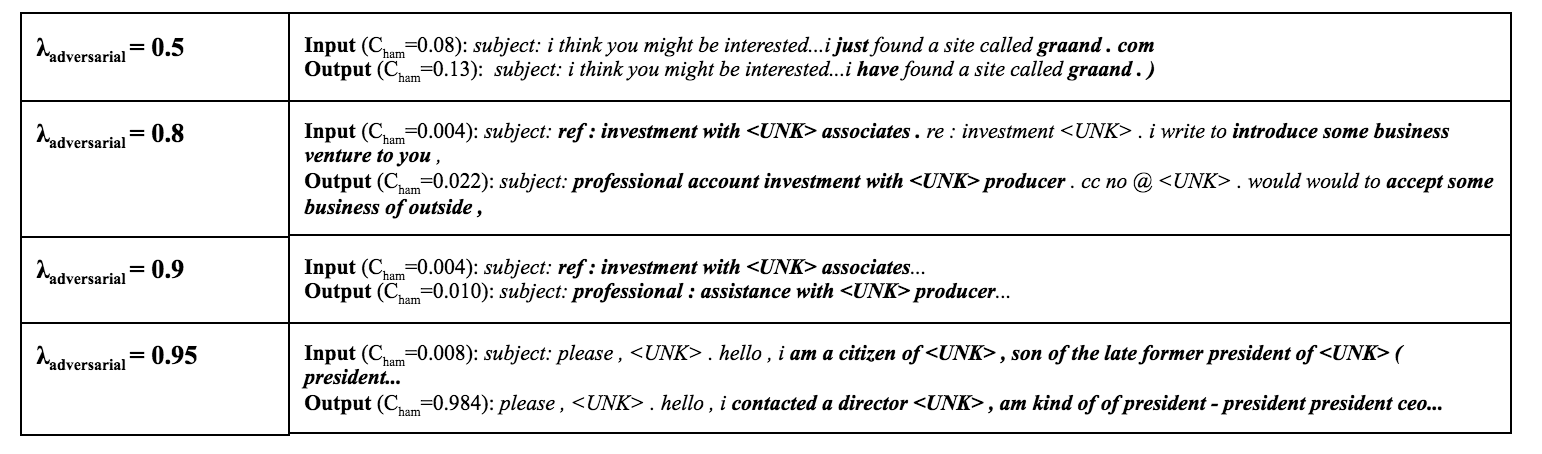}
  \caption{Excerpts from sampled original spam inputs and rewritten adversarial outputs after training on (n=171) low-confidence spam examples.}
\end{figure}

After training on the full set of n=1717 randomly sampled, previously unseen spam examples, we observe that one significant challenge in generating adversarial text examples may be the task simplicity, even for the relatively simple target classifier. The majority of spam and ham examples from the Enron dataset simply do not look alike to a bag-of-words classifier, making it quite difficult for the generator to discover enough semantically similar perturbations to fool the target classifier. In this section, we therefore draw on ideas from teacher forcing to attempt to avoid or prolong training instability by instead training within an easier, less reward-sparse domain. In particular, we order all n=1717 spam examples by their discriminator confidence, and select the 10\% (n=171) of spam examples with the \textit{lowest discriminator confidence} in the spam label for training. Intuitively, training the generator on these examples allows the generator to observe a larger quantity of tokens that may be less strongly associated with the spam label, guiding exploration within an easier domain.

Sampled excerpts from training on these low-confidence spam examples are shown in Figure 5. Ultimately, we find that training in this domain is still prone to the same training instability, leading to generator mode collapse after several training epochs. However, a subjective evaluation of generator outputs prior to mode collapse, as seen in Table 2, suggests that the generator appears to learn more complex, semantically-meaningful token substitutions, which more strongly resemble actual ham tokens and token sequences from the Enron dataset. These results suggest that pre-training on lower-confidence examples offers a promising pre-training regimen for generating adversarial examples - for example, a staggered training regimen on increasingly difficult examples may better drive exploration in discovering adversarial, semantically-meaningful token substitutions.

\section{Discussion}
\textbf{Summary.}  In this work, we motivated and introduced DANCin SEQ2SEQ, a framework for generating adversarial text examples targeting black-box text classifiers. Drawing on REINFORCE formulations that adapt GAN-style training to the discrete text generation domain, we recast adversarial example generation as a reinforcement learning task, and introduce an algorithm that rewards a generator for producing semantically similar perturbations to input text sequences that increase the probability of misclassification by the targeted model. 

Using an actual text classification task - spam filtering, trained on the real-world Enron spam dataset - we find that our algorithm produces encouraging, but highly preliminary: in particular, training results suggest that the generator does in fact learn to identify and remove highly spam-correlated tokens based on largely black-box interactions with the target classifier. Even more promisingly, in many cases, the generator learns adversarial token substitutions that increase the probability of misclassification while still preserving human semantic similarity.

\textbf{Limitations and Future Work.} While initial experiments suggest that GAN-style training offers a promising framework for adversarial example generation, we also find that - as with previous research on similar GAN-style learning regimens - training is highly unstable, and prone to divergence. We anticipate, however, that many strategies employed for reinforcement-learning-based text generation could yield similar stability improvements: for example, Li. et. al. suggest applying a penalty for repeated tokens to avoid the common mode collapse case, and offer a reward for every generation step (REGS) formulation that allows the generator to receive feedback even before the end of the full sequence generation episode  
\citep{li2017adversarial}. Future exploration could also consider reintroducing a true GAN-style discriminator, updated during training, to distinguish between human and machine generated examples. This could further reward higher quality generated adversarial examples.

As this work evaluates the DANCin SEQ2SEQ framework on the real-world but fairly limited binary spam-ham classification task, much work remains to further evaluate the algorithm on more challenging tasks, and against other target model architectures. Interestingly, we observe that even the learned adversarial examples that do not preserve semantic similarity, but instead exploit the bag-of-words nature of the Naive Bayes classifier used here, reveal fundamental assumptions made by the target classifier - without access to the model training details, these adversarial examples intuitively suggest how the target model approaches classification, and thus how it might fail. Future experimentation on more complex classification tasks and model architectures could yield similar interpretable insights. As with adversarial example generation in the image domain, this algorithm therefore offers a useful framework for understanding important target model vulnerabilities and limitations.

Overall, we believe that this work offers a promising approach towards understanding practical attacks on text-based classifiers, a high impact but previously under-explored domain. We hope that generating and understanding adversarial text examples will motivate further work in building more robust, nuanced, and well-defended text classification models.

\subsubsection*{Acknowledgments}

Many thanks to Will Monroe for his crackerjack adversarial text generation advice and expertise, and for sharing an alarming series of articles about 3D-printed turtles misclassified as guns. Additional thanks to Animesh Garg and Charles Lu for guiding discussion on the theme of “how can I fool classifiers with reinforcement learning”, and finally, to George Fei, for encouraging all kinds of fruitless verbal gymnastics around the terrible pun that constitutes this paper and method title.

\bibliography{cs332}
\bibliographystyle{plainnat}

\end{document}